\documentclass[10pt,twocolumn,letterpaper]{article}

\pdfoutput=1
\usepackage{cvpr}
\usepackage{times}
\usepackage{epsfig}
\usepackage{graphicx}
\usepackage{amsmath}
\usepackage{amssymb}
\usepackage{multirow}
\usepackage{booktabs}
\usepackage{comment}
\usepackage[labelfont=bf]{caption}
\usepackage[font={small}]{caption}
\usepackage{textcomp}

\usepackage[pagebackref=true,breaklinks=true,letterpaper=true,colorlinks,bookmarks=false]{hyperref}

\cvprfinalcopy 

\def\cvprPaperID{779} 

\ifcvprfinal\fi
\begin{document}

\title{Unsupervised Domain Adaptation for Object Detection\\ via Cross-Domain Semi-Supervised Learning}


\author{
{Fuxun Yu$^\dagger$, ~~Di Wang$^\ddagger$, ~~Yinpeng Chen$^\ddagger$, ~~Nikolaos Karianakis$^\ddagger$, ~~Tong Shen$^\ddagger$, ~~Pei Yu$^\ddagger$,} \\
{Dimitrios Lymberopoulos$^\ddagger$, Sidi Lu$^*$, Weisong Shi$^*$, Xiang Chen$^\dagger$} \\
{$^\dagger$ George Mason University,  $^\ddagger$ Microsoft}, $^*$	Wayne State University\\
}


\maketitle

\begin{abstract}
Current state-of-the-art object detectors can have significant performance drop when deployed in the wild due to domain gaps with training data.
Unsupervised Domain Adaptation (UDA) is a promising approach to adapt models for new domains/environments without any expensive label cost.
However,  without ground truth labels, most prior works on UDA for object detection tasks can only perform coarse image-level and/or feature-level adaptation by using adversarial learning methods.
In this work, we show that such adversarial-based methods can only reduce domain style gap, but cannot address the domain content distribution gap that are shown to be important for object detectors.
To overcome this limitation, we propose the Cross-Domain Semi-Supervised Learning (CDSSL) framework by leveraging high quality pseudo labels to learn better representations from target domain directly. 
To enable SSL for cross-domain object detection, we propose fine-grained domain transfer, progressive-confidence-based label sharpening and imbalanced sampling strategy to address two challenges: (i) non-identical distribution between source and target domain data, (ii) error amplification/accumulation due to noisy pseudo labeling on target domain.
	Experiment results show that our proposed approach consistently achieves new state-of-the-art performance (2.2\% - 9.5\% better than prior best work on mAP) under various domain adaptation scenarios. 
	Code will be released at $-link-$. 
\end{abstract}
\vspace{-3mm}
\section{Introduction}
\label{sec:intro}

Recently, deep learning has achieved superior performance for various vision tasks such as classification, object detection and semantic segmentation~\cite{alexnet, fastrcnn, maskrcnn, fcnn}. 
However, real world deployment environments can be highly uncertain and non-stationary, which can cause domain gaps including appearance/style difference (\textit{e.g.}, lighting, brightness) and content distribution mismatch (\textit{e.g.}, object density) with training data, resulting significantly reduced model performance~\cite{dabb, residual_transfer, adda, domain_confusion}. 
Traditional practice is to collect new training data to retrain models, but labeling these data can be very costly and time-consuming, especially for complex tasks such as object detection and segmentation~\cite{sim10k, label}. 
As a result, Unsupervised Domain Adaptation (UDA) using labeled source domain and unlabeled target domain has become a promising approach~\cite{dabb}.

\begin{figure}[!t]
	\centering
	\includegraphics[width=3.3in]{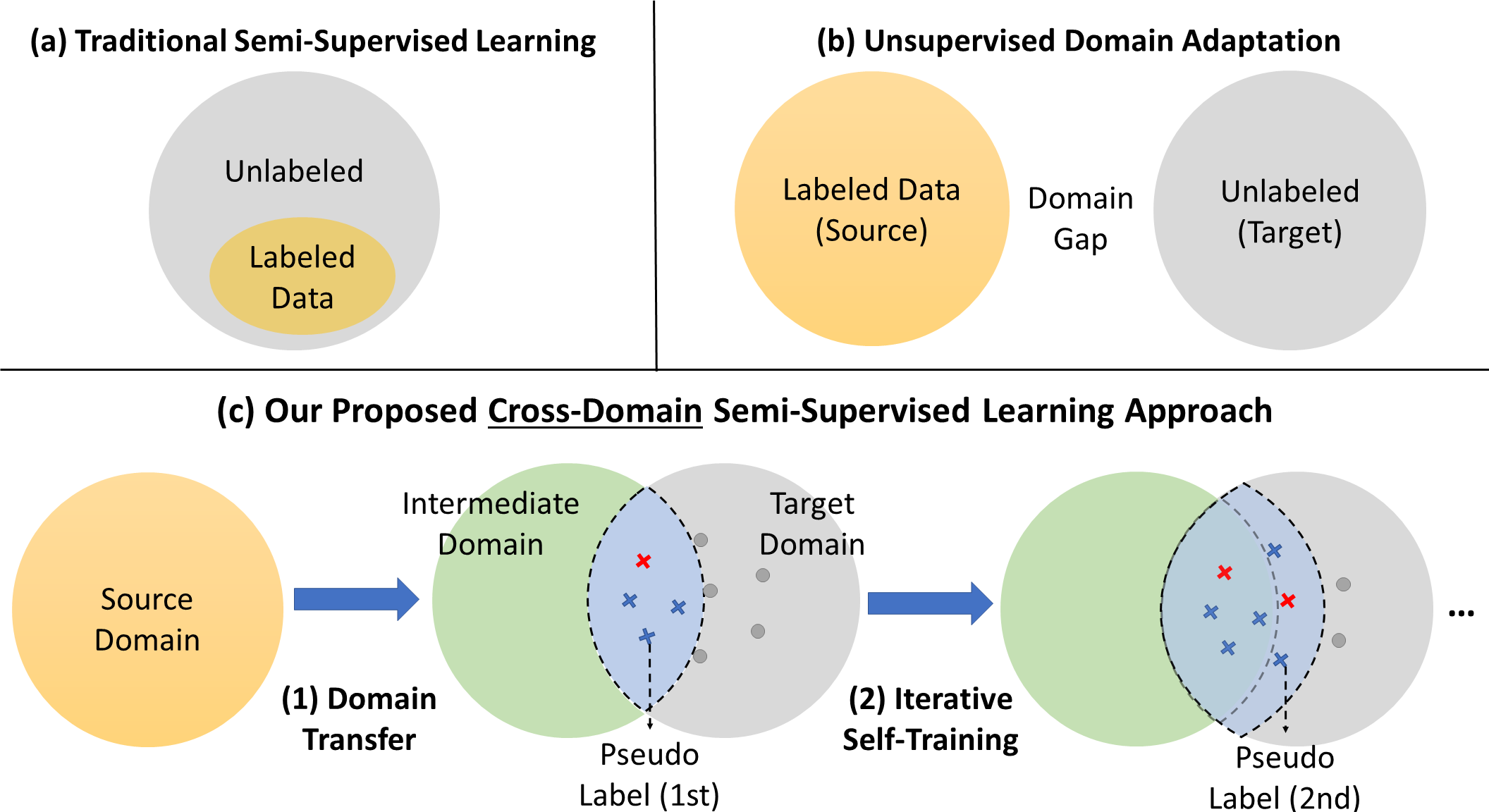}
	\vspace{-4mm}
        \caption{\textbf{Top:} Comparison of SSL and UDA settings. In UDA, labeled and unlabeled data do not come from the same distribution which is different from traditional SSL assumption.  
	\textbf{Bottom:} Our proposed CDSSL framework: 
	(1) Domain transfer to generate an intermediate domain to reduce the distribution gap;
	(2) Iterative Self-training to improve pseudo label quality on target domain.
        }
	\label{fig:page1}
	\vspace{-5mm}
\end{figure}

Prior UDA works have proposed many methods including metric-based discrepancy minimization, manifold/subspace alignment, domain mapping and domain-invariant feature learning, \textit{etc}~\cite{metric, subspace, cycada, dabb}. 	
	Most of these works target at image classification, where coarse-grained alignment by aligning features globally between source and target domain may suffice. 
	On the other hand, UDA for object detection may require not only coarse-grained/global alignment but also fine-grained/local alignment such as local object context, object density, \textit{etc.} that have been shown to be important for object detectors~\cite{scda, strong_weak}. 
	The state-of-the-art UDA works for object detectors ~\cite{wild, scda, strong_weak, multi_level} mostly use adversarial feature alignment methods to align features of unpaired source and target domain images. 
	However, feature alignment cannot distinguish objects belonging to different classes or in different contexts.
	Thus, it is prone to misalignment, \textit{e.g.}, misaligning features of foreground \textit{v.s.} background, features of different classes, or features of multiple \textit{v.s.} single object. 
Moreover, due to the lack of target-domain labels, adversarial methods can only work at a coarse image or feature level and may not be able to align any content mismatch between domains.

In this work, we take a different approach to reduce domain gaps for object detectors by leveraging Semi-Supervised Learning (SSL) concept: \textit{Self-Training}~\cite{self_train_2005}.  Self-Training method utilizes a portion of labeled data to generate pseudo labels for unlabeled data from the same distribution and combines both for training ~\cite{hard_label}.
Although such pseudo labels are not as accurate as ground-truth (GT) labels, they bring several benefits for object detections compared to prior coarse-grained feature alignment: 
(i) the detection model learns object features directly from pseudo labeled boxes, which is more fine-grained and with less ambiguity,
(ii) the overall pseudo labels on the target domain provide a way to approximate the real data distribution, and
(iii) by improving pseudo label quality,  better representations can be learned to improve detection performance.

However, as shown in Figure~\ref{fig:page1}a, traditional SSL assumes that labeled and unlabeled data come from the same distribution~\cite{goodfellow, ssl_overview, entropy_minimization, vda_overview}.  
	Consequently, directly applying SSL methods into our problem usually cannot work well due to two main challenges:
	(1) \textit{Non-identical distribution}: as shown in Figure~\ref{fig:page1}b, the source and target domain data typically come from different distributions with a domain gap;
	(2) \textit{Error amplification/accumulation}: due to the domain gap, the generated pseudo labels on target domain often contain many annotation errors. These erroneous labels will reinforce themselves during iterative SSL and generate more erroneous labels which significantly degrade object detection performance.

To address these challenges, we propose the Cross-Domain Semi-Supervised Learning (CDSSL) framework for object detectors  as shown in Figure~\ref{fig:page1}c. The CDSSL framework consists of two major steps:
(1) \textit{Domain Transfer.} To reduce non-identical data distribution gap, an intermediate domain is generated by transforming source domain image style to match with target domain. A detection model is trained on this intermediate domain as an initial pseudo label annotator on target domain.
(2) \textit{Iterative Self-Training.} To improve pseudo label quality, two key components are proposed: (i) imbalanced mini-batch sampling by over-sampling source domain samples and under-sampling target domain samples to reduce the impact of erroneous pseudo labels in each training iteration; (ii) progressive-confidence-based label sharpening to reduce pseudo-label error amplification effect by sharpening most confident predictions as pseudo/hard labels iteratively.

We evaluate our proposed CDSSL framework on several object detector adaptation benchmarks, including \textit{synthetic-to-real}, \textit{cross-camera}, and \textit{normal-to-foggy} that represent different degree of domain gaps. Our approach performs consistently better results than prior best work by 2.2\% - 9.5\% mAP, achieving the new state-of-the-art object detection domain adaptation performance.
Through detailed analysis, we demonstrate the importance of each key component of our CDSSL towards domain adaptation improvements and show how it can help object detection tasks.

\section{Related Work}
\label{sec:related}

\paragraph{Unsupervised Domain Adaptation.}

State-of-the-art deep neural networks often face significant performance drop due to the changing environments. 
	Therefore, many unsupervised domain adaptation (UDA) techniques are proposed, \textit{e.g.} MMD distance minimization~\cite{mmd}, sub-space alignment~\cite{subspace}, \textit{etc.}
	Recently, an adversarial learning based method with gradient reverse layer for feature-invariant learning achieves great performance for classification adaptation problems~\cite{dabb}.
	Many works also generalize it into other vision tasks including semantic segmentation~\cite{segmentation_da1, segmentation_da2}, object detection~\cite{wild}, \textit{etc.}
However,  UDA has different implications on image classification and object detection: (i) each input image for classification often only contain a single object of a category, while multiple objects of different categories may exist in a single image for object detections; (ii) detection task consists of both classification and localization which may require more than coarse-grained alignment as for classification.

\begin{figure*}[!tb]
	\centering
	\includegraphics[width=6.9in]{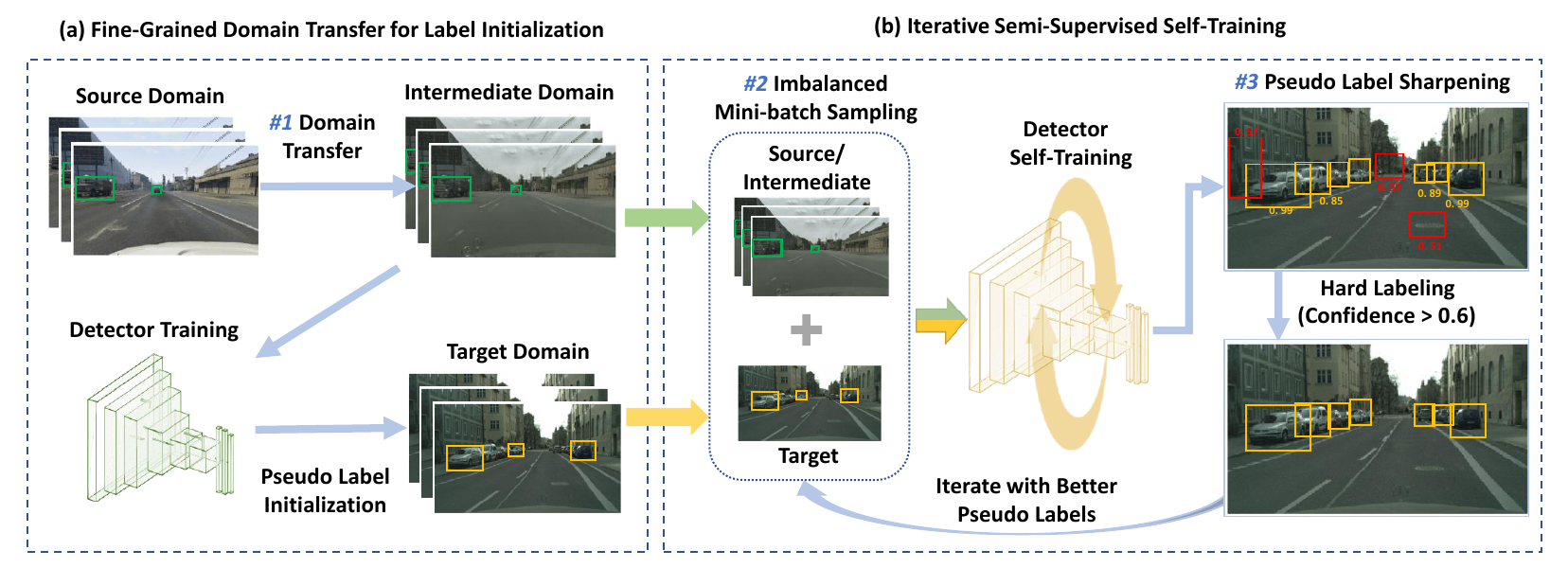}
	\vspace{-5mm}
	\caption{Cross-Domain Semi-Supervised Learning (CDSSL) Framework Overview. 
	(a) $^\textbf{\#1}$Fine-grained domain transfer first transfers the source to an intermediate domain to initialize high-quality pseudo labels on the target domain. 
	(b) Then we conduct iterative semi-supervised self-training with combined source/intermediate/target domain data.
	Specifically, $^\textbf{\#2}$imbalanced mini-batch sampling is proposed to prevent pseudo label errors from overwhelming the mini-batch during the detector self-training. 
	After each round, the new model is adopted for $^\textbf{\#3}$pseudo label sharpening on the target domain.
	This process can iterate multi-times by replacing with better pseudo labels.
	}
	\label{fig:overview}
	\vspace{-3mm}
\end{figure*}

\vspace{-4mm}
\paragraph{Domain Adaptation for Detection.}
Most prior works on domain adaptation for object detections used adversarial learning based methods to learn invariant features between domains.
	For example, Chen \textit{et al.}~\cite{wild} first use adversarial learning methods to align the image-level and instance-level features.
	Following that, Zhu \textit{et al.}~\cite{scda}  propose a region-level alignment method to target at a middle granularity.
	Some other adversarial methods have also targeted at different levels with weighting strategies~\cite{strong_weak, multi_level, progressive}.
These methods can well solve the style gaps between different domains.
	But due to the challenges like multi-objects, multi-classes, and varied-locations, they may cause feature misalignment.
	Moreover, these methods mostly focus on coarse-granularity feature alignment and do not account for content distribution gap (\textit{e.g.}, different object density distribution) which are important for object detections.

\vspace{-4mm}
\paragraph{Semi-Supervised Learning.}

Semi-supervised learning (SSL) has become a promising approach utilizing both labeled data and large amount of unlabeled data to not only reduce the labeling cost but also improve the model performance. Many SSL methods have been proposed including EM algorithm~\cite{em}, self-training~\cite{self_train_2005}, co-training~\cite{co_train}, transduction methods~\cite{transductive}, as well as deep learning based SSL ones~\cite{deep_semi1, deep_semi2, deep_semi3}. 
On the other hand, SSL does not always work well, as discussed in~\cite{goodfellow, entropy_minimization}. 
One common assumption for SSL to work is the \textit{cluster assumption}, \textit{i.e.}, the labeled and unlabeled data tend to form a same or similar distributed cluster~\cite{goodfellow, ssl_overview}. 
However, for domain adaptation, this assumption often cannot be satisfied due to the large domain gaps. 
Another common assumption in SSL is the \textit{low density prior}, \textit{i.e.}, the classifier in real-world settings favors a low-density separation manifold between classes~\cite{low_density_theory}.
	Such prior motivates the commonly-used hard labeling, entropy minimization method, \textit{etc}~\cite{hard_label, entropy_minimization}.
	In this work, when applying SSL in UDA for object detection, we assume that our problem setting satisfies the low density prior since the foreground/background and different classes in our datasets (\textit{i.e.}, street scene datasets) are highly distinguishable. 

\section{Methodology}
\label{sec:method}

In this work, we address the unsupervised domain adaptation between the labeled source domain $(X_s, ~Y_s)$ and the unlabeled target domain $(X_t, ~?)$, where the $?$ denotes that ground-truth annotation $Y_t$ in target domain is unavailable.

\vspace{-3mm}
\paragraph{CDSSL Framework Overview.}

The CDSSL framework consists of two major steps as shown in Fig.~\ref{fig:overview}: \textbf{(a)} Because of the cross-domain setting, the labeled data and unlabeled data does not meet 
the identical distribution assumption for semi-supervised learning.
	This can cause the subsequent self-training to fail in some complex tasks.
	Therefore, we first conduct \textit{domain transfer} to get an intermediate domain with smaller gaps from target domain.
	The initial pseudo label annotator will then be trained on this intermediate domain to generate higher-quality initial pseudo labels. 
	This step only needs to be performed once in order to launch essential next self-training step;
\textbf{(b)} We then run \textit{semi-supervised self-training} for detector training by combing the source/intermediate domain data (with GT labels) and the target domain data (with pseudo labels) together.
	To prevent the pseudo label errors from overwhelming the training process, we conduct \textit{imbalanced mini-batch sampling} during training process.
	After each self-training round, the model can often yield better performance on target domain. 
	Therefore, we utilize the better model for \textit{confidence-based hard labeling} to get better pseudo labels.
	The self-training process could \textit{iterate multiple times} till the model performance stops improving because of the continuing and irreversible error accumulation from erroneous pseudo labels.

\subsection{Detection-Oriented Domain Transfer}

The aim of domain transfer is to transfer source-domain data into an intermediate domain, which is closer to the target domain distribution.
	One common solution is to conduct domain style translation in image level by cyclegan~\cite{cyclegan}.
	In this section, we show that naive cyclegan translation can only achieve suboptimal translation results for detection tasks.
	Then we propose to optimize it by receptive filed restriction for detection-oriented style translation.

\vspace{-4mm}
\paragraph{Cross-Domain Style Translation.}

In our setting, we assume the source domain and target domain images have two different styles. 
	Therefore, we adopt the popular image translation model cyclegan~\cite{cyclegan} for our cross-domain style transfer.
	For cyclegan training, one source-domain image will be translated by generators twice: from source to target domain; and then translated back, ensuring the cycle-consistency between original and reconstructed images.
	Meanwhile, two discriminators will discriminate the generated images from real images to ensure the image styles match the corresponding domain styles. 

However, original cyclegan design does not target at any specific end tasks. 
	As a result, when applying to detection images, vanilla cyclegan often not only translates the styles, but also changes the objects and backgrounds. 
	This is prohibitive for detection task since every object has a matched label in the dataset. 
	To solve this problem, we optimize it as \textit{fine-grained cyclegan} to conduct detection-oriented style translation specialized for object detection tasks.

\begin{figure}[!tb]
	\centering
	\includegraphics[width=3.3in]{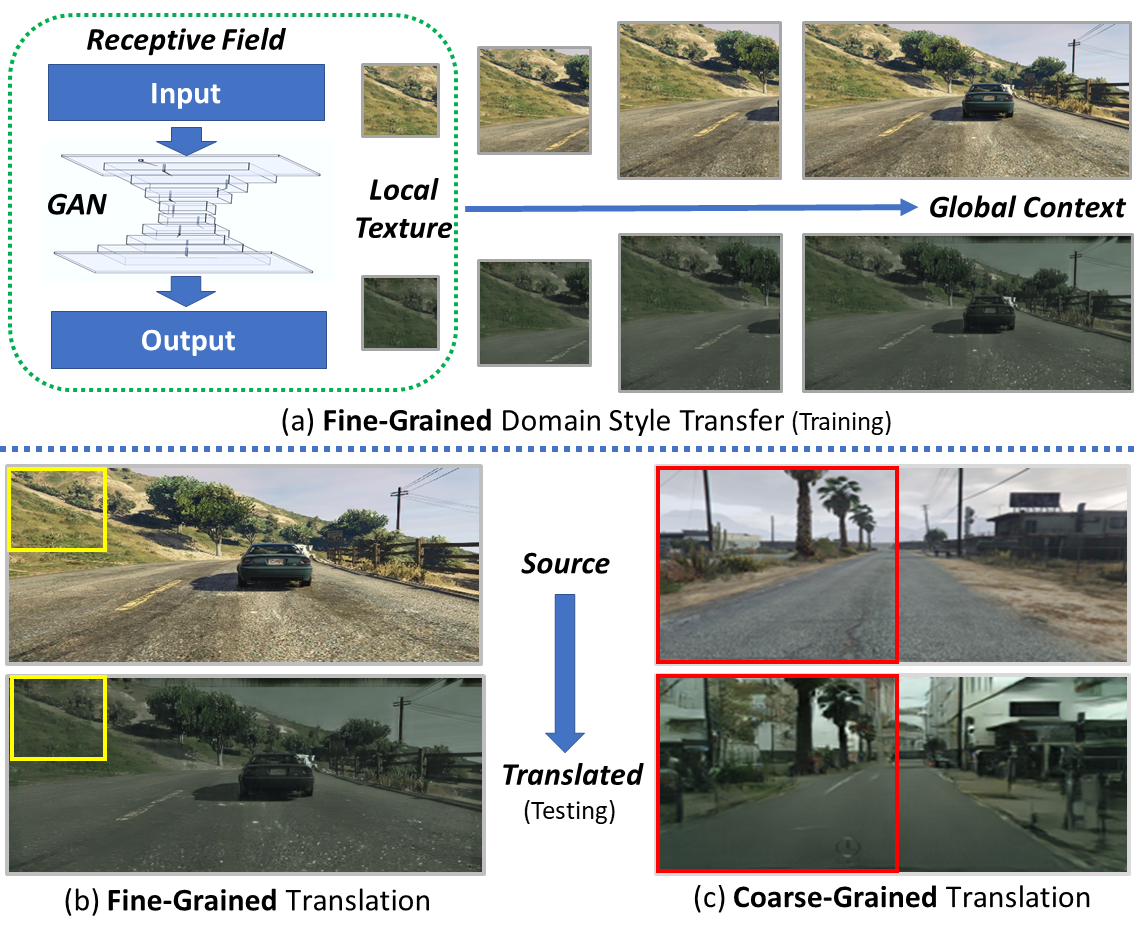}
	\vspace{-5mm}
	\caption{CycleGAN with large receptive field tends to translate/change the objects and its background context, which is suboptimal for object detection adaptation tasks.}
	\label{fig:patch-gan}
	\vspace{-5mm}
\end{figure}

\vspace{-3mm}
\paragraph{Receptive Field Restriction for Fine Granularity.} 

As we discussed, a detection-oriented image translator should not change foregrounds and backgrounds (\textit{i.e.}, global content) but focus on the low-level styles (\textit{e.g.}, local texture, color and brightness).
To do so, we propose a simple yet effective method: imposing restrictions on the receptive field of cyclegan model.
As illustrated in Fig.~\ref{fig:patch-gan} (a), in training process, we restrict the input patch size of the generator and discriminator so that the models can only ``see'' the local texture difference without knowing the global contents or major contexts. 
Thus, cyclegan model can only learn to translate and discriminate the size-restricted patch style.
During testing, by using the fully-convolutional model structure, the generator can be applied on the full-size image for fine-grained style translation.
Figure~\ref{fig:patch-gan} (b) and (c) compare the translation results between fine-/coarse-grained translation. 
For fine-grained translation, only \textit{local texture} in the source domain are changed to the translated domain, while for coarse-grained translation, 
\textit{global contents} such as the background of source domain are dramatically changed to match target domain.
 As shown in Section~\ref{sec:analysis}, such translation results often achieve suboptimal performance compared to our fine-grained translation results.

\subsection{Semi-Supervised Self-Training}

\paragraph{Pseudo Label Initialization via Intermediate Domain.}

After the domain transfer, we can get an intermediate domain $X_m$ that is closer to the target domain, \textit{i.e.}, $X_s \rightarrow X_m \rightarrow X_t$.
	As we mentioned earlier, the basic self-training setting directly uses the source domain $(X_s, ~Y_s)$ to train the initial annotator.
	But in our setting, if source/target domain gap $X_s\rightarrow X_t$ is too large, model trained on source domain can often produce many erroneous pseudo labels. 
	This can cause the following self-training to fail in complex tasks such as object detection.
	To reduce the amount of erroneous pseudo labels, we conduct annotator training on the intermediate domain $(X_m, ~Y_s)$. Since the gap between $X_m \rightarrow X_t$ is smaller, the annotator can thus generate more accurate pseudo labels $Y_t^{'}$ on target domain $X_t$. 
	This initialization step is only executed once,  and then our self-training can iteratively boost pseudo label quality to improve object detection performance on the target domain.

\vspace{-4mm}
\paragraph{Imbalanced Sampling for Error Mitigation.}

Annotated by non-perfect labelers, the pseudo labels in target domain $(X_t, ~Y_t^{'})$ inevitably contain some errors. 
	Without any precautions, this can greatly hinder the detector's learning process. 
	For example, in the RPN training with fore/back-ground classification loss:
	\begin{equation}
	\small
	L_t=-[~y_i^{'}\log(p_i) + (1-y_i^{'})\log(1-p_i)~], ~~i \in t,
	\end{equation}
	\normalsize
	$p_i$ is the model predicted foreground probability.
	If the pseudo label $y_i^{'}=1$ for some background anchor, the wrong classification loss $L_t$ would mislead the model to predict the $p_i$ for this and similar background anchors' confidence as 1.
	The model can then become biased and more similar incorrect pseudo labels will appear later.
	This is called \textit{Error Accumulation/Amplification}, another major problem that prevents the self-training from getting better performance.

To mitigate the bias of inaccurate target-domain loss $L_t$, we combine source/intermediate\footnote{In practice, combining which one (source or intermediate) depends on different datasets (which yields some slight difference, see Supp. materials). Later on, we unify the terms as `source'/target domain combination.} and target domain data to conduct imbalanced sampling in each data mini-batch.
	Since the source-domain samples have ground-truth labels $(X_s, ~Y_s)$, the loss $L_s$ should be accurate. 
	Therefore, we statistically under-sample the target domain samples but over-sample the source domain samples during training. 
	In this case, the loss of each training batch is composed of:
	\vspace{-1mm}
	\begin{equation}
	\small
	Loss = \sum_{i=0}^{s} {L_i} + \sum_{j=0}^{t} {L_j}, ~~\textbf{s.t.}~~ s > t.
	\vspace{-1mm}
	\end{equation}
	\normalsize
	where $s$ and $t$ denote the sampling amounts of source domain and target domain data, respectively.
	In case that the pseudo labels contain errors, the learning loss of source-domain samples $L_s$ can still produce reliable gradients for each training batch.
	Therefore, such imbalanced sampling strategy can provide training stabilization and error-correction effects, as shown in Section~\ref{sec:analysis}. 

\vspace{-4mm}
\paragraph{Confidence-based Hard Pseudo Labeling.}

At the end of each self-training round, the model's performance on target domain could become better than the previous annotator's.
	Therefore, we can update the pseudo labels on the target domain by applying the better model on the target domain.
	To control the label quality, we only choose the most confident predictions above the confidence threshold as pseudo labels.
	And based on the low density assumption~\cite{low_density_theory}, we sharpen the soft predicted probability distribution into hard labels to learn more confident representations:
	\vspace{-1mm}
	\begin{equation}
	\small
	\begin{split}
		\medmuskip=-2mu
		y_i = \left\{
		\begin{aligned}
			 &\operatorname*{arg\,max}_i(p_i), &\text{$\max(p_i) > $ conf};\\
			 &0, &\text{Otherwise}.
		\end{aligned}
		\right.
	\label{eq:mask}
	\end{split}
	\vspace{-2mm}
\end{equation}
\normalsize
	In the case of two classes, this will binarize the prediction $y_i$ based on confidence threshold $conf$.
	In multi-class situations, the pseudo label would be chosen if the highest confidence score $\max(p_i)$ exceeds the confidence threshold.
	Otherwise, it will still be considered as background.

During iterative self-training rounds, some error boxes' confidence will be reinforced and similar wrong predictions can appear later with higher confidence.
	To avoid including too many errors, we progressively increase the confidence threshold in later rounds to maintain our label quality. 

\section{Experimental Evaluation}

\subsection{Experiments Setup}

For experiments setup, we follow the same settings as in~\cite{scda}. 
	We use Faster-RCNN with VGG16 backbone as our object detection network. 
	Three benchmarking domain adaptation scenarios are evaluated, namely Synthetic to Real (\textit{Sim2City}), Cross Camera (\textit{Kitti2City}) and Normal to Foggy (\textit{City2Foggy}).
	For the image size, previous works mainly use two settings: 512 pixels or 600 pixels as image's shorter side. 
	Experiments using higher resolution (600 pixels) usually achieve better performance.
	For fair comparison, we evaluate our framework under both settings.
	For the evaluation metric, we report the mean average precision (mAP) at IoU threshold 0.5. 
	Code will be available.

\begin{table}[tb!]
\renewcommand\arraystretch{0.8}
\centering
\caption{SIM10K to Cityscapes Adaptation Performance. Number $^{512}$ denotes images are resized with 512 pixels as shorter side, so does $^{600}$. For fair comparison, we compare both settings.}
\vspace{-2mm}
\setlength{\tabcolsep}{6.5mm}{
\small
\begin{tabular}{l|c}
\toprule[0.9pt]
\multicolumn{1}{c|}{Methods} & \textit{Car AP} \\ \midrule[0.9pt]
$^{512}$Baseline {[}non-adapt{]} & 33.0 \\ \hline
$^{512}$Faster-rcnn in the wild~\cite{wild} & 39.0 \\ \hline
$^{512}$SCDA~\cite{scda} & 43.0 \\ \hline
$^{512}$Ours w/ Self-Train (ST) & 40.6 \\ \hline
$^{512}$Ours w/ Domain-Transfer (DT) & 48.1 \\ \hline
$^{512}$Ours w/ (DT + ST) & \textbf{49.0} \\ \hline
$^{512}$Oracle Performance & {61.6} \\ \midrule[0.9pt] \midrule[0.9pt]
$^{600}$Strong-Weak~\cite{strong_weak} & 42.3 \\ \hline
$^{600}$Multi-Level~\cite{multi_level} & 42.8 \\ \hline
$^{600}$Ours w/ Domain-Transfer (DT) & 50.8 \\ \hline
$^{600}$Ours w/ (DT + ST) & \textbf{52.3} \\ \hline
$^{600}$Oracle Performance& {62.7} \\ \bottomrule[0.9pt]
\end{tabular}}
\vspace{-4mm}
\label{table:sim2city}
\end{table}
\normalsize

\begin{table*}[!t]
\renewcommand\arraystretch{0.7}
\centering
\vspace{-5mm}
\caption{Multi-Class Cityscapes to Foggy-Cityscapes Adaptation Performance}
\vspace{-3mm}
\setlength{\tabcolsep}{3.2mm}{
\small
\begin{tabular}{l|cccccccc|l}
\toprule[0.9pt]
\multicolumn{1}{c|}{Methods} & \multicolumn{1}{c}{Person} & \multicolumn{1}{c}{Rider} & \multicolumn{1}{c}{Car} & \multicolumn{1}{c}{Truck} & \multicolumn{1}{c}{Bus} & \multicolumn{1}{c}{Train} & \multicolumn{1}{c}{Motor} & \multicolumn{1}{c|}{Bicycle} & \multicolumn{1}{c}{mAP} \\ \midrule[0.9pt]
$^{512}$Baseline [non-adapt] & 29.7 & 32.2 & 44.6 & 16.2 & 27.0 & 9.1 & 20.7 & 29.7 & 26.2 \\ \hline
$^{512}$Faster-rcnn in the wild.~\cite{wild} & 25.0 & 31.0 & 40.5 & 22.1 & 35.3 & 20.2 & 20.0 & 27.1 & 27.6 \\ \hline
$^{512}$SCDA~\cite{scda} & 33.5 & 38.0 & 48.5 & 26.5 & 39.0 & 23.3 & 28.0 & 33.6 & 33.8 \\ \hline \hline
$^{512}$Ours w/ Self-Train (ST) & 30.0 & 33.1 & 46.5 & \underline{11.2} & \underline{17.7} & \underline{5.0} & 21.1 & 30.6 & \underline{24.4} \\ \hline
$^{512}$Ours w/ Domain-Transfer (DT) & 33.6 & 39.2 & 52.2 & 22.4 & 41.3 & 27.4 & 24.8 & 34.1 & 34.3 \\ \hline
$^{512}$Ours w/ (DT + ST) & {33.9} & 38.7 & 52.1 & 26.3 & 43.4 & \textbf{32.9} & 27.5 & 35.5 & \textbf{36.3} \\ \hline
$^{512}$Oracle Performance & 40.7 & 44.7 & 61.9 & 28.2 & 51.3 & 33.0 & 31.4 & 40.9 & 41.5 \\ \midrule[0.9pt] \midrule[0.9pt]
$^{600}$Strong-Weak~\cite{strong_weak} & 29.9 & 42.3 & 43.5 & 24.5 & 36.2 & 32.6 & 30.0 & 35.3 & 34.3 \\ \hline 
$^{600}$Multi-Level~\cite{multi_level} & 33.2 & \textbf{44.2} & 44.8 & \textbf{28.2} & 41.8 & 28.7 & {30.5} & {36.5} & 36.0 \\ \hline \hline
$^{600}$Ours w/ Domain-Transfer (DT) & 36.5 & 41.7 & 54.6 & 22.6 & 40.7 & 25.3 & 29.3 & 36.9 & 35.9 \\ \hline
$^{600}$Ours w/ (DT + ST) & \textbf{38.2} & 42.1 & \textbf{55.6} & 25.9 & \textbf{43.5} & {27.6} & \textbf{33.5} & \textbf{39.2} & \textbf{38.2} \\ \hline
$^{600}$Oracle Performance & 42.7 & 49.2 & 63.4 & 35.8 & 53.1 & 22.7 & 33.5 & 39.7 & 42.5 \\ \bottomrule[0.9pt]
\end{tabular}}
\vspace{-3mm}
\label{table:city2foggy}
\end{table*}
\normalsize

\begin{table}[!tb]
\renewcommand\arraystretch{0.8}
\centering
\caption{KITTI to Cityscapes Adaptation Performance}
\vspace{-3mm}
\setlength{\tabcolsep}{6.5mm}{
\small
\begin{tabular}{l|c}
\toprule[0.9pt]
\multicolumn{1}{c|}{Methods} & \textit{Car AP} \\ \midrule[0.9pt]
$^{512}$Baseline {[}non-adapt{]} & 36.4 \\ \hline
$^{512}$Faster-rcnn in the wild~\cite{wild} & 38.5 \\ \hline
$^{512}$SCDA~\cite{scda} & 42.5 \\ \hline
$^{512}$Ours w/ Self-Train (ST) & 42.6 \\ \hline
$^{512}$Ours w/ Domain-Transfer (DT) & 41.4 \\ \hline
$^{512}$Ours w/ (DT + ST) & \textbf{45.2} \\ \hline
$^{512}$Oracle Performance & {61.6} \\ \midrule[0.9pt] \midrule[0.9pt]
$^{600}$Ours w/ Domain-Transfer (DT) & 42.8 \\ \hline
$^{600}$Ours w/ (DT + ST) & \textbf{46.4} \\ \hline
$^{600}$Oracle Performance& {62.7} \\ \bottomrule[0.9pt]
\end{tabular}}
\vspace{-5mm}
\label{table:kitti2city}
\end{table}
\normalsize

\subsection{Domain Adaptation Performance}

\paragraph{Synthetic to Real Adaptation.}

We use SIM10K $\rightarrow$ Cityscapes as the domain pairs for synthetic to real adaptation scenario. 	
SIM10K is a synthetic dataset generated by GTA-V game engine~\cite{sim10k}, and Cityscapes consists of images of real street scenes taken at different cities~\cite{cityscapes}. 
Since SIM10K dataset only contains \textit{car} annotation, we focus on \textit{car} detection task. 
The car detection results of our approaches are compared with the baseline and state-of-the-art methods~\cite{wild, scda, strong_weak, multi_level} as shown in Table~\ref{table:sim2city}.
Baseline represents applying source-domain trained car detection model directly on the target domain validation set without any domain adaptation. 
The  Self-Train setting (ST) denotes only applying our iterative self-training but without the domain transfer step, and vice versa for the fine-grained domain transfer (DT) setting. 
Oracle represents car detection performance trained on labeled target domain.
%
Compared to Baseline, applying ST and DT alone improves \textit{car} AP by +7.6\% and +15.1\%, respectively.
	By combining both methods (DT+ST), the result boosts up to 49.0\% AP.
	Compared to prior SOTA works, our approach achieves the current best performance, +6.0\% and +9.5\% better than prior best work in 512 and 600 pixel resolutions, respectively.

\vspace{-4mm}
\paragraph{Cross Camera Adaptation.}

We also evaluate the performance of our framework in cross camera scenarios using KITTI $\rightarrow$ Cityscapes~\cite{kitti}.
	As per setting in \cite{wild, scda}, only \textit{car} annotation is used for training and evaluation.
	The overall results are shown in Table~\ref{table:kitti2city}.
Compared to Baseline, ST and DT alone improves AP by +6.2\% and +5.0\%, respectively. In constrast with synthetic-to-real adaptation, DT achieves relatively smaller improvement in cross-camera adaptation and we hypothesize that the domain style gap is smaller between two real-world street scene datasets.
	This causes the improvement potential for domain transfer to be smaller, but in turn can make self-training work better. 
	Combining both (ST+DT) brings +8.8\% improvement over Baseline, which outperforms the previous best result~\cite{scda} by 2.7\%.

\vspace{-4mm}
\paragraph{Multi-Class Normal to Foggy Adaptation.}

Multi-class adaptation scenarios are more challenging than two-classes fore/background scenarios.
	In this experiment, we evaluate our framework on Cityscapes $\rightarrow$ Foggy-Cityscapes~\cite{foggy}. 
	Following~\cite{wild, scda}, eight categories are used for training and evaluation.
	As shown in Table~\ref{table:city2foggy},
 our approach achieves the best performance under both resolution settings, achieving +2.5\% and +2.2\% mAP gain compared to prior SOTA performance. 
	Furthermore, our performance has nearly achieved the oracle model performance for several classes. 
	This indicates we have nearly closed the domain gaps, demonstrating the effectiveness of our CDSSL framework.

Interestingly, the self-training only (ST) approach performs worse (24.4\%) than Baseline (26.2\%) in such a multi-class adaptation scenario. 
	This is different from previous single-class results where ST only usually improves over Baseline.
	By examining each class, we find that the performance drop lies in three classes (\textit{Truck, Bus} and \textit{Train}).
	Due to the original large domain gap, the baseline performance for these classes is significantly worse than other classes.
	This causes the initial pseudo label quality for these classes to be very low, containing a lot of errors. 
	During self-training, such low-quality pseudo labels will `teach' the following self-trained models to produce even worse performance: \textit{16.2 $\rightarrow$ 11.2, 27.0 $\rightarrow$ 17.7, 9.1 $\rightarrow$ 5.0}. 
By contrast, after our domain transfer step, these three classes' performance become on pair with other classes due to the reduced domain gap. 
In addition, combining with self-training (ST+DT) turns to benefit even more, achieving +10.1\% mAP than Baseline, and +2.0\% than DT. 
	Specifically, for the three classes with bad initial performance, the improvements are even more significant: \textit{16.2 $\rightarrow$ 26.3, 27.0 $\rightarrow$ 43.4, 9.1 $\rightarrow$ 32.9}.
	Such evident results support our hypothesis: \textit{when source-target gaps are too large, the self-training can hardly work and may even hurt performance compared to baseline approach, demonstrating the necessity of domain transfer in our overall framework.}
\section{Ablation Study for Design Modules}
\label{sec:analysis}

\begin{figure}[!tb]
	\centering
	\includegraphics[width=3.3in]{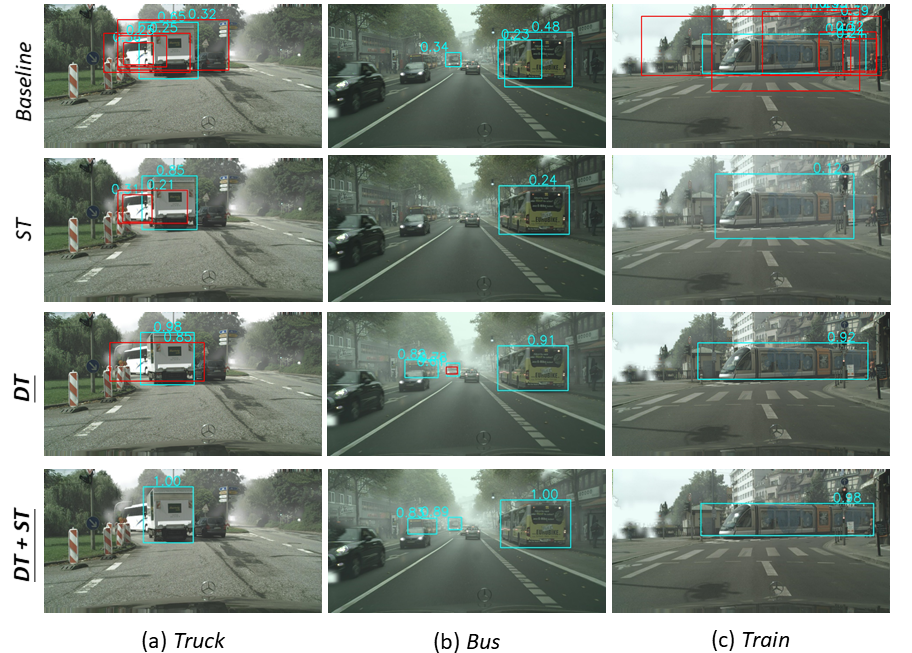}
	\vspace{-6mm}
	\caption{Pseudo Label Quality Visualization on three classes in Cityscapes-Foggy. Labelers are models from baseline, self-training only, domain transfer, and domain transfer + self-training.}
	\label{fig:label_quality}
	\vspace{-5mm}
\end{figure}

\paragraph{Impact of CDSSL on Pseudo Label Quality.} 
In Figure~\ref{fig:label_quality}, we visualize the effectiveness of our CDSSL framework in improving pseudo label quality.
	Due to the large domain gap, Baseline model generates very low-quality pseudo labels, \textit{i.e.}, erroneous predictions with very low confidence-score.
	Directly trained on these Baseline pseudo labels, the self-trained (\textit{ST}) model becomes even worse at detecting these classes, as shown in Table.~\ref{table:city2foggy}.
	By contrast, with domain transfer, the DT model significantly improves prediction accuracy with high confidence. Therefore, high-quality pseudo labels (\textit{i.e.}, higher precision and recall with GT) are generated,  and Table.~\ref{table:city2foggy} shows the quantitative improvement.
	Furthermore, with our proposed hard label sharpening, the subsequent self-trained model (\textit{ST+DT}) can receive the highest-quality supervision, thus achieving the best performance among all settings.

\begin{table}[tb!]
\renewcommand\arraystretch{1.0}
\centering
\caption{Effectiveness of Fine-grained Domain Transfer. 
Note *: The max. patch size is set to 360x360 for Kitti2City since its image max. height is 375.}
\vspace{-2mm}
\small
\setlength{\tabcolsep}{3.6mm}{
\begin{tabular}{c|ccc}
\toprule[1pt]
\multicolumn{1}{c|}{Patch-Size} & \textit{Sim2City} & \textit{Kitti2City} & \textit{City2Foggy} \\ \midrule[1pt]
Non-Adapt & 33.0 & 36.4 & 26.2\\ \hline
512x512 & 42.4 & 40.5* & 30.0 \\ \hline
256x256 & 44.3 & 40.0 & 30.6 \\ \hline
128x128 & \textbf{48.1} & \textbf{41.4} & \textbf{34.3} \\ \bottomrule[1pt]
\end{tabular}}
\vspace{-2mm}
\label{table:patchsize}
\end{table}
\normalsize

\vspace{-3mm}
\paragraph{Impact of Fine-Grained Domain Transfer.} 
To study the impact of receptive filed restriction, we trained cyclegan models under three patch-size settings: $128^2$, $256^2$, $512^2$.
	These models are then used to translate the source domain data for training object detection model using Faster-RCNN.
	As shown in Table~\ref{table:patchsize}:  (i) Domain adaption with different patch sizes all give better for object detection performance (mAP) than no adaptation (Non-Adapt) which shows the importance reducing domain style gap; and (ii) As patch size reduces, domain adaptation for object detection performance (mAP) consistently improves (\textit{e.g.}, 5.7\% difference between $512^2$ and $128^2$ in \textit{Sim2City}) as image contents are better maintained with finer-granularity translation.

\begin{table}[!tb]
\renewcommand\arraystretch{1.0}
\centering
\caption{Imbalanced Sampling Impact. ``Source : Target'' denotes the sampling ratio. Results are drawn from 1st self-training round.}
\vspace{-2mm}
\small
\setlength{\tabcolsep}{3.1mm}{
\begin{tabular}{ccccc}
\toprule[1pt]
Source~:~Target               & 0:4 & 1:3  & 2:2  & 3:1  \\ \midrule[1pt]
\textit{Sim2City} {[}conf=0.6{]}   & \underline{46.2} & 47.0 & 48.2 & \textbf{49.0} \\ \hline 
\textit{Kitti2City} {[}conf=0.5{]} & 41.5 & \underline{40.6} & 42.6 & \textbf{43.5} \\ \hline
\textit{Kitti2City} {[}conf=0.6{]} & \underline{40.8} & 41.0 & 41.8 & \textbf{44.3} \\ \hline
\textit{Kitti2City} {[}conf=0.7{]} & \underline{40.5} & 41.4 & 41.3 & \textbf{42.0} \\ \hline 
\textit{City2Foggy} {[}conf=0.6{]} & \underline{32.5} & 35.1 & 35.6 & \textbf{35.7} \\ \bottomrule[1pt]
\end{tabular}}
\vspace{-4mm}
\label{table:mini-batch}
\end{table}
\normalsize

\vspace{-3mm}
\paragraph{Impact of Imbalanced Mini-batch Sampling.}

Pseudo labels generated by annotators contain errors, which can hurt the model learning process. 
	As a remedy, imbalanced sampling aims to mitigate the errors' influence during model training.
	To verify its efficacy, we set different sampling ratios during training and evaluate the corresponding training performance.
	As shown in Table.~\ref{table:mini-batch}, the source and target sampling ratio, S:T=0:4 \textit{i.e.}, no source data being sampled gives the worst performance for most settings.
	By adding one or more source-domain samples in each mini-batch, the performance consistently improves under almost all settings and achieves the best at 3:1 ratio. 
		Note that setting the ratio of S:T to 4:0 deduces to training purely on source domain, which is our Baseline setting.
Such results demonstrate that our imbalanced sampling strategy indeed makes the training process more robust to label errors. 
	Our hypothesis is the ground-truth labels in source domain can still provide roughly correct gradient direction in each training iteration. 
	Therefore, with higher sampling ratio of source-domain samples, the model can converge to a better minimum in terms of overall detection loss.

\vspace{-4mm}
\paragraph{Impact of Progressive Confidence Thresholding.}

\begin{figure}[!tb]
	\centering
	\includegraphics[width=3.3in]{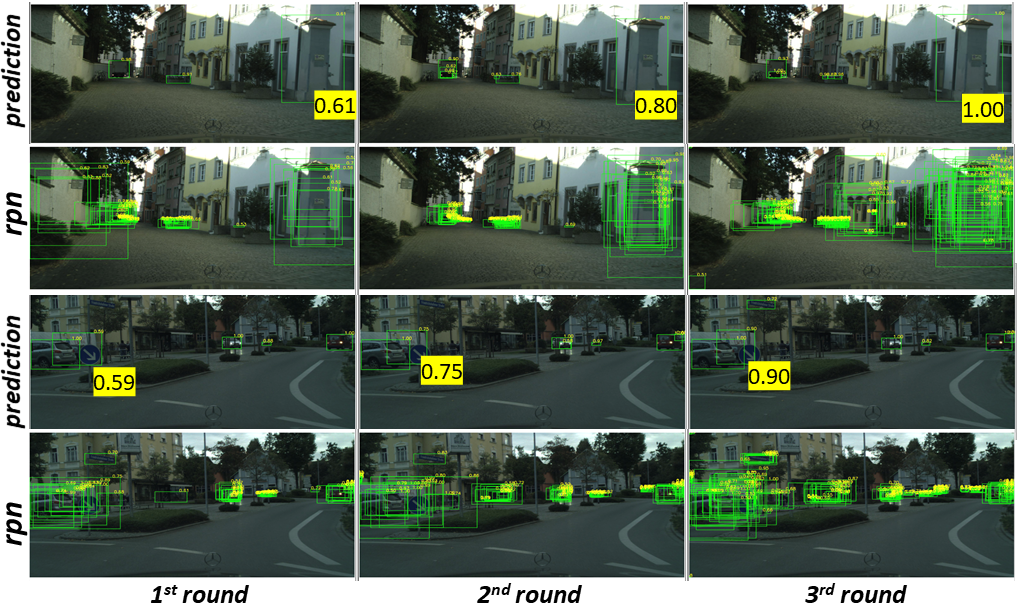}
	\vspace{-5mm}
	\caption{During iterative self-training, some wrong boxes' confidence can continually increase, \textit{i.e.}, the \textit{error accumulation}.
	Because of this, the rpn proposals also become very low-quality. }
	\vspace{-3mm}
	\label{fig:error_acc}
\end{figure}

\begin{table}[tb!]
\renewcommand\arraystretch{1.0}
\centering
\caption{Progressive Conf-Thresholding. 
Every new self-training round (one column) uses the best annotator from the last column.}
\vspace{-2mm}
\small
\setlength{\tabcolsep}{2.4mm}{
\begin{tabular}{c|ccc|ccc}
\toprule[1pt]
                 & \multicolumn{3}{c|}{\textit{Kitti2City}}                & \multicolumn{3}{c}{\textit{City2Foggy}}       \\ \hline
Conf-Thres & 1st           & 2nd           & 3rd           & 1st           & 2nd           & 3rd  \\ \midrule[1pt]
0.5              & 43.5          & 43.8          & -             & 34.1          & -             & -    \\ \hline
0.6              & \textbf{44.3} & 43.7          & 43.8          & \textbf{35.7} & 35.1          & 35.0 \\ \hline
0.7              & 42.0          & 44.5          & 43.9          & 33.2          & 35.2          & 34.9 \\ \hline
0.8              & 42.2          & \textbf{45.2} & 43.9          & 33.1          & \textbf{36.3} & 34.4 \\ \hline
0.9              & 32.5          & 43.3          & \textbf{44.7} & -             & 35.9          & \textbf{35.0} \\ \bottomrule[1pt]
\end{tabular}}
\label{table:conf-thres}
\vspace{-4mm}
\end{table}
\normalsize

Even with imbalanced sampling strategy, the \textit{Error Accumulation} problem can still happen. 
	As shown in Figure~\ref{fig:error_acc}, erroneous pseudo labels can iteratively reinforce themselves and increase their confidence scores.
	Because of such wrong supervision, the RPN proposals in later self-training rounds also become highly biased to error regions.
	Such substantial label errors are the major problems that prevent iterative self-training evolving with better performance. 

As a mitigation, Table.~\ref{table:conf-thres} demonstrates the necessity of progressive confidence thresholding:
	In both adaptation scenarios, only setting higher thresholds in later iterations can help maintain the performance increasing trend. 
	And after three iterations, the performance on both datasets saturates and cannot further improve due to the large amounts of label errors.
Currently, error accumulation in self-training is still an open problem~\cite{error_accum1, error_accum2, error_accum3}. 
	And it is more complicated for object detection task as the label error can consist of both classification and regression errors. 
We plan to explore better error mitigation approaches in future work.

\vspace{-2mm}
\section{Discussion}
\vspace{-1mm}

\paragraph{Where does CDSSL improvement come from?}
\label{sec:why_help}
Given that object detection needs to optimize both classification and regression/localization tasks, it is important to understand the impact of each individual task under different sources of supervisions (source domain GT label vs. target domain pseudo label) on the adaptation performance. 
Specifically, we set binary masks on classification loss and regression loss for source domain GT labels and target domain pseudo labels, respectively. 
This enables backpropagating the loss of either regression head or classification head during iterative self-training.
\begin{table}[!tb]
\renewcommand\arraystretch{1.0}
\centering
\caption{Analysis of performance improvement in CDSSL. The evaluated datasets are Kitti2City. Pseudo\_City means training on Cityscapes dataset which are pseudo labeled by DT model.
}
\vspace{-2mm}
\small
\setlength{\tabcolsep}{5.1mm}{
\begin{tabular}{l|l}
\toprule[1pt]
\multicolumn{1}{c|}{Train Settings} & AP \\ \midrule[1pt]
KITTI [baseline] & 36.4 \\ \hline
Domain Transfer (DT) & 41.4 \\ \hline
Pseudo\_City (labeled by DT model) & 40.8 \\ \midrule[1pt]
KITTI + Pseudo\_City (+ only regression) & 31.3 \\ \hline
KITTI + Pseudo\_City (+ only classification) & 43.5 \\ \hline
KITTI + Pseudo\_City (+ both) & \textbf{44.3} \\ \bottomrule[1pt]
KITTI (+ only regression) + Pseudo\_City  & \textbf{44.4} \\ \hline
KITTI (+ only classification) + Pseudo\_City  & 43.5 \\ \hline
KITTI (+ both) + Pseudo\_City & 44.3 \\ \bottomrule[1pt]
\end{tabular}}
\label{table:source-improve}
\vspace{-5mm}
\end{table}
\normalsize

From Table~\ref{table:source-improve}, we have several interesting observations: 
	(1) For the target-domain pseudo labels, their classification loss supervision seems to be more important than regression one:
	Enabling classification loss or classification + regression losses dramatically improves performance (+7.1$\sim$7.9\%), while the regression loss alone degrades the model performance than baseline (-5.1\%). 
	This could be because the pseudo labels' box coordinates are mostly inaccurate, but they still partially cover the right objects, and thus enables the model to learn better feature representations from the target domain and provides beneficial classification supervision; 
	(2) For the source-domain GT labels, both loss components benefit the performance while the regression part seems to be the major benefit:
	The regression and classification supervision of source domain can improve +3.6\% and +2.7\% AP over Pseudo\_City setting, which is trained without source-domain labels;
	(3) As expected, Pseudo\_City model trained on only pseudo labels achieves even worse performance than the pseudo label annotator (-0.6\%) due to the error amplification/accumulation effect of erroneous pseudo labels.
	This further demonstrates the importance of combining source-domain GT labels in the self-training process, which can provide training-stabilization/error-correction effects.

These empirical results show that our CDSSL framework may achieve a synergy effect by benefiting from the \textit{source domain GT-label regression supervision} and the \textit{target domain pseudo-label classification supervision}. 

\vspace{-3mm}
\paragraph{Improvement beyond Style Gap.}

\begin{figure}[!tb]
	\centering
	\includegraphics[width=3.3in]{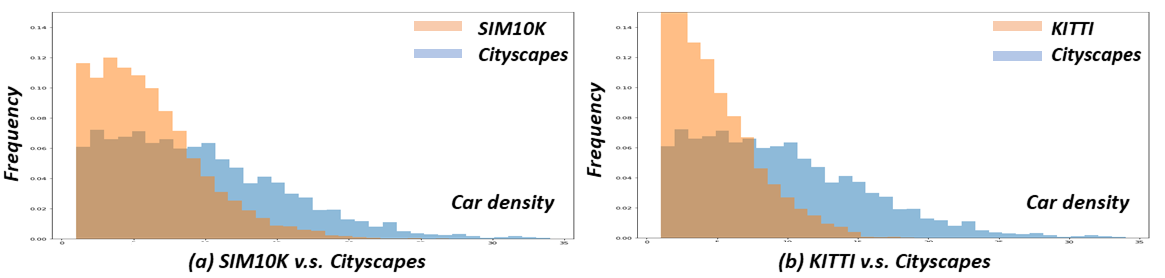}
	\vspace{-6mm}
	\caption{Car density distribution comparison, showing that \textit{Content Distribution Gap} can exist in different detection domains.}
	\label{fig:label_distribution}
	\vspace{-3mm}
\end{figure}

In object detection, detectors inherently need to handle {multi-scale}, {multi-class} objects with {varied densities} in a single image.
	This can add new dimension of gaps beyond \textit{style gap}, which we call \textit{content distribution gap} such as object density as shown in Figure~\ref{fig:label_distribution}.
Here we empirically show that previous feature alignment works mainly address the style gap, but our approach can further reduce the content distribution gap.
We compare our approach with  \textit{Frcnn in the wild}~\cite{wild}, one of the representative feature-level adversarial learning methods. 

\begin{table}[tb!]
\renewcommand\arraystretch{0.9}
\centering
\caption{SIM10K to Cityscapes Adaptation on MaskRCNN. Images of full resolution are used for training and evaluation.
}
\vspace{-2mm}
\small
\setlength{\tabcolsep}{6.0mm}{
\begin{tabular}{l|c}
\toprule[1pt]
\multicolumn{1}{c|}{Methods} & \textit{Car AP} \\ \midrule[1pt]
Baseline {[}non-adapt{]} & 45.4 \\ \hline
Frcnn in the wild~\cite{wild} & 52.5 \\ \midrule[1pt]
Fine-Grained Domain Trans. (DT) & 62.8 \\ \hline
DT + Frcnn in the wild~\cite{wild} & 63.1 (+0.3) \\ \hline
DT + ST-1st Iter & 66.1 (+3.3) \\ \hline
DT + ST-2nd Iter & 68.1 (+5.3) \\ \hline
DT + ST-3rd Iter & \textbf{69.8} (+7.0) \\ \bottomrule[1pt]
\end{tabular}}
\vspace{-4mm}
\label{table:maskrcnn}
\end{table}
\normalsize
As Table.~\ref{table:maskrcnn} shows,  \textit{Frcnn} approach can achieve 52.5 AP with 7.1\% improvement over Baseline with no domain adaption. 
However, by first applying fine-grained domain transfer and use this style translated dataset as the new source domain (62.8 AP), \textit{Frcnn} can barely achieve improvement (only +0.3\%).
We hypothesize that the style translation has already reduced the style gap, and consequently the feature-alignment method cannot achieve further gain since that they are mostly targeting at the same style gap but at different levels (i.e., pixel level vs. feature level).
In contrast, our CDSSL approach by leveraging iterative self-training (ST) can give an additional 7\% improvement on top of style translation, far more than the adversarial-based techniques, such as \textit{Frcnn}. 
This implies that beyond reducing the image-level or feature-level gap, our CDSSL framework may also help reduce the content distribution gap in a semi-supervised way, suggesting a new direction in addressing UDA for object detection.

\vspace{-2mm}
\section{Conclusion}
\vspace{-1mm}

In this work, we propose CDSSL: a cross-domain semi-supervised learning framework to address the unsupervised domain adaptation for object detection. Specifically, we conduct domain transfer to meet the distribution assumption of SSL and then lauch our iterative self-training.
	Optimizations including imbalanced sampling and confidence-based label sharpening are proposed to mitigate the error accumulation. 
	Experiments demonstrate the effectiveness of our framework, which consistently outperforms previous SOTA by 2.2\% - 9.5\% in various adaptation scenarios.

{\small
\bibliographystyle{ieee_fullname}
\bibliography{egbib.bib}

\begin{thebibliography}{10}\itemsep=-1pt

\bibitem{ssl_overview}
Olivier Chapelle, Bernhard Scholkopf, and Alexander Zien.
\newblock Semi-supervised learning (chapelle, o. et al., eds.; 2006)[book
  reviews].
\newblock {\em IEEE Transactions on Neural Networks}, 20(3):542--542, 2009.

\bibitem{low_density_theory}
Olivier Chapelle and Alexander Zien.
\newblock Semi-supervised classification by low density separation.
\newblock In {\em AISTATS}, volume 2005, pages 57--64. Citeseer, 2005.

\bibitem{co_train}
Minmin Chen, Kilian~Q Weinberger, and John Blitzer.
\newblock Co-training for domain adaptation.
\newblock In {\em Advances in neural information processing systems}, pages
  2456--2464, 2011.

\bibitem{wild}
Yuhua Chen, Wen Li, Christos Sakaridis, Dengxin Dai, and Luc Van~Gool.
\newblock Domain adaptive faster r-cnn for object detection in the wild.
\newblock In {\em Proceedings of the IEEE conference on computer vision and
  pattern recognition}, pages 3339--3348, 2018.

\bibitem{transductive}
Ronan Collobert, Fabian Sinz, Jason Weston, and L{\'e}on Bottou.
\newblock Large scale transductive svms.
\newblock {\em Journal of Machine Learning Research}, 7(Aug):1687--1712, 2006.

\bibitem{cityscapes}
Marius Cordts, Mohamed Omran, Sebastian Ramos, Timo Rehfeld, Markus Enzweiler,
  Rodrigo Benenson, Uwe Franke, Stefan Roth, and Bernt Schiele.
\newblock The cityscapes dataset for semantic urban scene understanding.
\newblock In {\em Proceedings of the IEEE conference on computer vision and
  pattern recognition}, pages 3213--3223, 2016.

\bibitem{subspace}
Basura Fernando, Amaury Habrard, Marc Sebban, and Tinne Tuytelaars.
\newblock Unsupervised visual domain adaptation using subspace alignment.
\newblock In {\em Proceedings of the IEEE international conference on computer
  vision}, pages 2960--2967, 2013.

\bibitem{dabb}
Yaroslav Ganin and Victor Lempitsky.
\newblock Unsupervised domain adaptation by backpropagation.
\newblock {\em arXiv preprint arXiv:1409.7495}, 2014.

\bibitem{kitti}
Andreas Geiger, Philip Lenz, Christoph Stiller, and Raquel Urtasun.
\newblock Vision meets robotics: The kitti dataset.
\newblock {\em The International Journal of Robotics Research},
  32(11):1231--1237, 2013.

\bibitem{metric}
Bo Geng, Dacheng Tao, and Chao Xu.
\newblock Daml: Domain adaptation metric learning.
\newblock {\em IEEE Transactions on Image Processing}, 20(10):2980--2989, 2011.

\bibitem{fastrcnn}
Ross Girshick.
\newblock Fast r-cnn.
\newblock In {\em Proceedings of the IEEE international conference on computer
  vision}, pages 1440--1448, 2015.

\bibitem{entropy_minimization}
Yves Grandvalet and Yoshua Bengio.
\newblock Semi-supervised learning by entropy minimization.
\newblock In {\em Advances in neural information processing systems}, pages
  529--536, 2005.

\bibitem{mmd}
Arthur Gretton, Karsten~M Borgwardt, Malte~J Rasch, Bernhard Sch{\"o}lkopf, and
  Alexander Smola.
\newblock A kernel two-sample test.
\newblock {\em Journal of Machine Learning Research}, 13(Mar):723--773, 2012.

\bibitem{error_accum3}
Bo Han, Quanming Yao, Xingrui Yu, Gang Niu, Miao Xu, Weihua Hu, Ivor Tsang, and
  Masashi Sugiyama.
\newblock Co-teaching: Robust training of deep neural networks with extremely
  noisy labels.
\newblock In {\em Advances in neural information processing systems}, pages
  8527--8537, 2018.

\bibitem{maskrcnn}
Kaiming He, Georgia Gkioxari, Piotr Doll{\'a}r, and Ross Girshick.
\newblock Mask r-cnn.
\newblock In {\em Proceedings of the IEEE international conference on computer
  vision}, pages 2961--2969, 2017.

\bibitem{cycada}
Judy Hoffman, Eric Tzeng, Taesung Park, Jun-Yan Zhu, Phillip Isola, Kate
  Saenko, Alexei~A Efros, and Trevor Darrell.
\newblock Cycada: Cycle-consistent adversarial domain adaptation.
\newblock {\em arXiv preprint arXiv:1711.03213}, 2017.

\bibitem{segmentation_da1}
Judy Hoffman, Dequan Wang, Fisher Yu, and Trevor Darrell.
\newblock Fcns in the wild: Pixel-level adversarial and constraint-based
  adaptation.
\newblock {\em arXiv preprint arXiv:1612.02649}, 2016.

\bibitem{progressive}
Han-Kai Hsu, Wei-Chih Hung, Hung-Yu Tseng, Chun-Han Yao, Yi-Hsuan Tsai, Maneesh
  Singh, Ming-Hsuan Yang, Wayne Treible, Philip Saponaro, Yi Liu, et~al.
\newblock Progressive domain adaptation for object detection.
\newblock In {\em Proceedings of the IEEE Conference on Computer Vision and
  Pattern Recognition Workshops}, pages 1--5, 2019.

\bibitem{error_accum1}
Wei Hu, Jianfeng Chen, and Yuzhong Qu.
\newblock A self-training approach for resolving object coreference on the
  semantic web.
\newblock In {\em Proceedings of the 20th international conference on World
  wide web}, pages 87--96. ACM, 2011.

\bibitem{sim10k}
Matthew Johnson-Roberson, Charles Barto, Rounak Mehta, Sharath~Nittur Sridhar,
  Karl Rosaen, and Ram Vasudevan.
\newblock Driving in the matrix: Can virtual worlds replace human-generated
  annotations for real world tasks?
\newblock In {\em 2017 IEEE International Conference on Robotics and Automation
  (ICRA)}, pages 746--753. IEEE, 2017.

\bibitem{deep_semi1}
Durk~P Kingma, Shakir Mohamed, Danilo~Jimenez Rezende, and Max Welling.
\newblock Semi-supervised learning with deep generative models.
\newblock In {\em Advances in neural information processing systems}, pages
  3581--3589, 2014.

\bibitem{alexnet}
Alex Krizhevsky, Ilya Sutskever, and Geoffrey~E Hinton.
\newblock Imagenet classification with deep convolutional neural networks.
\newblock In {\em Advances in neural information processing systems}, pages
  1097--1105, 2012.

\bibitem{hard_label}
Dong-Hyun Lee.
\newblock Pseudo-label: The simple and efficient semi-supervised learning
  method for deep neural networks.
\newblock In {\em Workshop on Challenges in Representation Learning, ICML},
  volume~3, page~2, 2013.

\bibitem{error_accum2}
Ming Li and Zhi-Hua Zhou.
\newblock Setred: Self-training with editing.
\newblock In {\em Pacific-Asia Conference on Knowledge Discovery and Data
  Mining}, pages 611--621. Springer, 2005.

\bibitem{fcnn}
Jonathan Long, Evan Shelhamer, and Trevor Darrell.
\newblock Fully convolutional networks for semantic segmentation.
\newblock In {\em Proceedings of the IEEE conference on computer vision and
  pattern recognition}, pages 3431--3440, 2015.

\bibitem{residual_transfer}
Mingsheng Long, Han Zhu, Jianmin Wang, and Michael~I Jordan.
\newblock Unsupervised domain adaptation with residual transfer networks.
\newblock In {\em Advances in Neural Information Processing Systems}, pages
  136--144, 2016.

\bibitem{self_train_2005}
David McClosky, Eugene Charniak, and Mark Johnson.
\newblock Effective self-training for parsing.
\newblock In {\em Proceedings of the main conference on human language
  technology conference of the North American Chapter of the Association of
  Computational Linguistics}, pages 152--159. Association for Computational
  Linguistics, 2006.

\bibitem{em}
Geoffrey McLachlan and Thriyambakam Krishnan.
\newblock {\em The EM algorithm and extensions}, volume 382.
\newblock John Wiley \& Sons, 2007.

\bibitem{goodfellow}
Avital Oliver, Augustus Odena, Colin~A Raffel, Ekin~Dogus Cubuk, and Ian
  Goodfellow.
\newblock Realistic evaluation of deep semi-supervised learning algorithms.
\newblock In {\em Advances in Neural Information Processing Systems}, pages
  3235--3246, 2018.

\bibitem{vda_overview}
Vishal~M Patel, Raghuraman Gopalan, Ruonan Li, and Rama Chellappa.
\newblock Visual domain adaptation: An overview of recent advances.
\newblock {\em IEEE Signal Processing Magazine}, 2014.

\bibitem{label}
Stephan~R Richter, Vibhav Vineet, Stefan Roth, and Vladlen Koltun.
\newblock Playing for data: Ground truth from computer games.
\newblock In {\em European conference on computer vision}, pages 102--118.
  Springer, 2016.

\bibitem{strong_weak}
Kuniaki Saito, Yoshitaka Ushiku, Tatsuya Harada, and Kate Saenko.
\newblock Strong-weak distribution alignment for adaptive object detection.
\newblock In {\em Proceedings of the IEEE Conference on Computer Vision and
  Pattern Recognition}, pages 6956--6965, 2019.

\bibitem{foggy}
Christos Sakaridis, Dengxin Dai, and Luc Van~Gool.
\newblock Semantic foggy scene understanding with synthetic data.
\newblock {\em International Journal of Computer Vision}, 126(9):973--992,
  2018.

\bibitem{deep_semi3}
Antti Tarvainen and Harri Valpola.
\newblock Mean teachers are better role models: Weight-averaged consistency
  targets improve semi-supervised deep learning results.
\newblock In {\em Advances in neural information processing systems}, pages
  1195--1204, 2017.

\bibitem{adda}
Eric Tzeng, Judy Hoffman, Kate Saenko, and Trevor Darrell.
\newblock Adversarial discriminative domain adaptation.
\newblock In {\em Proceedings of the IEEE Conference on Computer Vision and
  Pattern Recognition}, pages 7167--7176, 2017.

\bibitem{domain_confusion}
Eric Tzeng, Judy Hoffman, Ning Zhang, Kate Saenko, and Trevor Darrell.
\newblock Deep domain confusion: Maximizing for domain invariance.
\newblock {\em arXiv preprint arXiv:1412.3474}, 2014.

\bibitem{deep_semi2}
Jason Weston, Fr{\'e}d{\'e}ric Ratle, Hossein Mobahi, and Ronan Collobert.
\newblock Deep learning via semi-supervised embedding.
\newblock In {\em Neural Networks: Tricks of the Trade}, pages 639--655.
  Springer, 2012.

\bibitem{multi_level}
Rongchang Xie, Fei Yu, Jiachao Wang, Yizhou Wang, and Li Zhang.
\newblock Multi-level domain adaptive learning for cross-domain detection.
\newblock In {\em Proceedings of the IEEE International Conference on Computer
  Vision Workshops}, pages 0--0, 2019.

\bibitem{segmentation_da2}
Yang Zhang, Philip David, and Boqing Gong.
\newblock Curriculum domain adaptation for semantic segmentation of urban
  scenes.
\newblock In {\em Proceedings of the IEEE International Conference on Computer
  Vision}, pages 2020--2030, 2017.

\bibitem{cyclegan}
Jun-Yan Zhu, Taesung Park, Phillip Isola, and Alexei~A Efros.
\newblock Unpaired image-to-image translation using cycle-consistent
  adversarial networks.
\newblock In {\em Proceedings of the IEEE international conference on computer
  vision}, pages 2223--2232, 2017.

\bibitem{scda}
Xinge Zhu, Jiangmiao Pang, Ceyuan Yang, Jianping Shi, and Dahua Lin.
\newblock Adapting object detectors via selective cross-domain alignment.
\newblock In {\em Proceedings of the IEEE Conference on Computer Vision and
  Pattern Recognition}, pages 687--696, 2019.

\end{thebibliography}
}

\newpage
\onecolumn

\section{Supplementary Materials}

The supplementary materials are organized as follows to supplement several details:

	Sec.~\ref{sec:dt} visualizes the style translation advantages of small receptive filed for cyclegan training in domain transfer.

	Sec.~\ref{sec:st} demonstrates the performance difference when combining source or intermediate domain in iterative self-training. 

	Sec.~\ref{sec:quali} qualitatively visualizes examples to show the adaptation performance improvement of our framework. 

	Sec.~\ref{sec:quati} quantitatively shows the performance improvements by analyzing the label quality distribution evolution.

\subsection{{{Domain Transfer}}: Results Visualization with Varied Patch Size}
\label{sec:dt}

\begin{figure*}[h]
	\centering
	\includegraphics[width=6.5in]{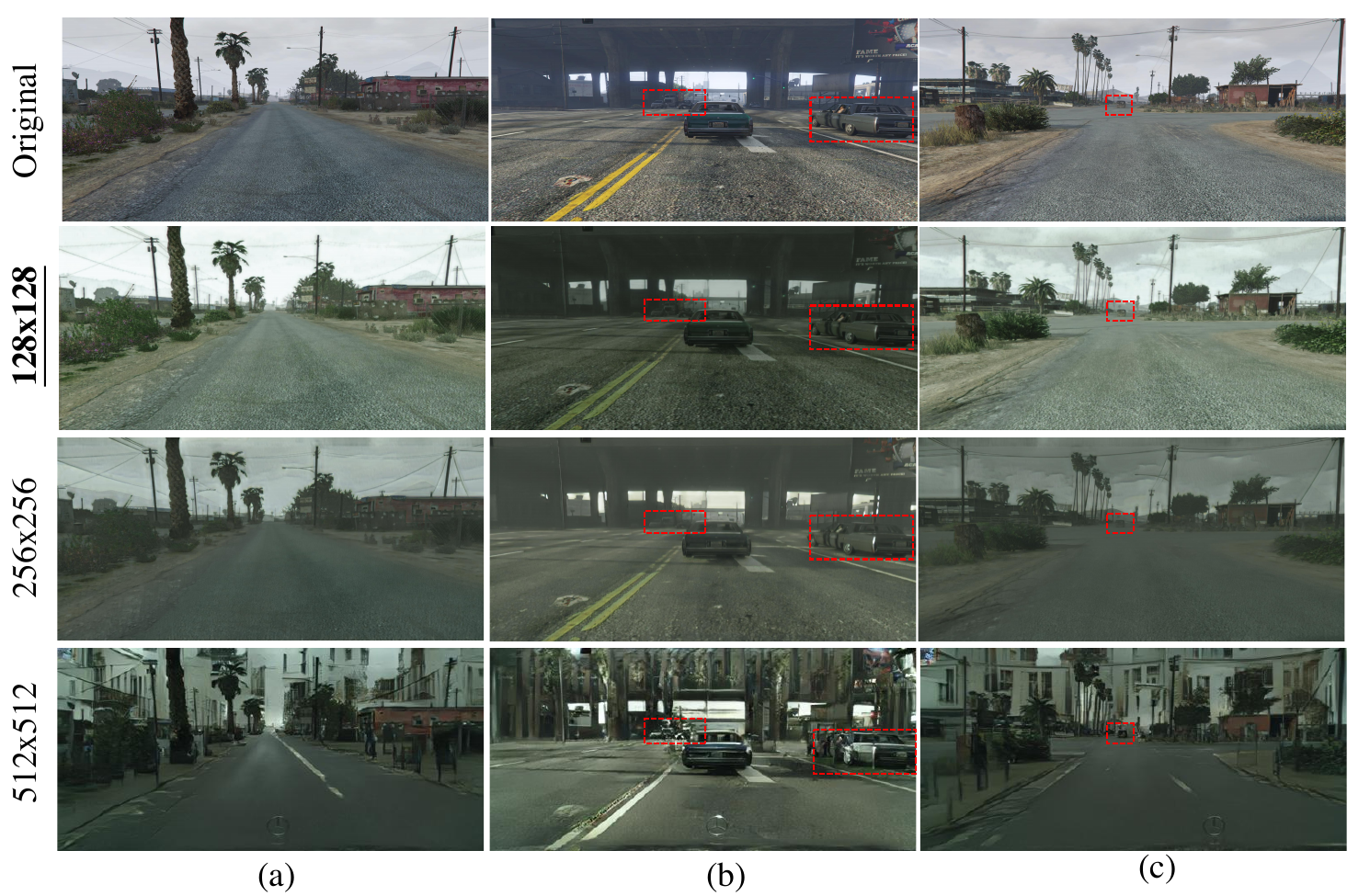}
	\vspace{-3mm}
	\caption{CycleGAN with large receptive filed generates artifacts and corrupts objects, which are harmful for object detection tasks.}
	\label{fig:patch-anal}
\end{figure*}

In this section, we show the benefits of restricting small receptive field during cyclegan training. 
	We visualize and compare the translation results of different cyclegan models in Fig.~\ref{fig:patch-anal}. 
	The first row (Original) is SIM10K images without translation.
	Other rows show the translated images by different cyclegans with corresponding receptive field sizes ($128^2 \rightarrow 512^2$). 
\begin{itemize}
	\item Qualitatively, we can observe that the translation results from small patch-size cyclegan models ($128^2$) \textbf{reserve the contents fairly well}. 
	Most edges and structures of cars are still distinguishable even for small-size cars. 
	
	\item By contrast, for the large patch-size cyclegan models ($512^2$), the translation results not only \textbf{dramatically changed the context} (\textit{e.g.}, the backgrounds in (a) and (c)), but also \textbf{modified the contents} (\textit{e.g.}, cars in red boxes).
\end{itemize}

Since the contents is the key of object detection, the generated destructive artifacts of large-patch cyclegan can mislead the model to learn incorrect features, and thus perform badly on the real target domain images.

\newpage

\subsection{{{Iterative Self-Training}}: Combination of Source or Intermediate Domain}
\label{sec:st}

In this section, we show the slight difference of combining source or intermediate domain with target domain during our self-training process.
	Note that, the pseudo labels on the target domain are always labeled by the annotator trained on the intermediate domain. 

The results are shown in Table.~\ref{table:combination}, which are drawn from the first iteration of self-training experiments.
	We compare two settings: \textit{source+target} and \textit{intermediate+target} on all three adaptation scenarios.
\begin{itemize}
	\item In \textit{Sim2City} and \textit{Kitti2City} adaptation, combining \textbf{source+target} yields slightly better performance. 

	\item However, on \textit{City2Foggy} adaptation, combining \textbf{intermediate+target} gives the better performance.
\end{itemize}
	
The performance variation of different combination lies in the range of $0.2\sim 1.3$ across all scenarios. 
	Our hypothesis is that on Sim2City and Kitti2City, since the source/target dataset have totally different contents, the domain style translation can still introduce some artifacts.
	As a result, combining source domain without artifacts may be slightly better. 
	But for City2Foggy, they have the same contents (cityscapes-foggy are generated by adding fogs on cityscapes), so the style transfer may introduce less artifacts.
	Therefore, combining intermediate domain can produce better results.

For our experiments in the following self-training iterations, we follow the best empirical results in each setting to choose the combination, \textit{i.e.}, we choose source+target for Sim2City and Kitti2City, while intermediate+target for City2Foggy. 

\begin{table*}[h]
\renewcommand\arraystretch{1.4}
\caption{The Performance Difference of Self-Training in terms of different combination. All experimental results are from the first iteration of self-training. Note: * denotes a shorter training setting (6 epochs), while all the others are using 12 training epochs. }
\setlength{\tabcolsep}{5.8mm}{
\begin{tabular}{l|cc|ccc|cc}
\hline \hline
\multicolumn{1}{c}{} & \multicolumn{2}{c}{\textit{Sim2City}} & \multicolumn{3}{c}{\textit{Kitti2City}} & \multicolumn{2}{c}{\textit{City2Foggy}} \\ \hline \hline
Combination & 0.7* & 0.7 & 0.5 & 0.6 & 0.7 & 0.5 & 0.6 \\ \hline
Source+Target & \textbf{0.482} & \textbf{0.490} & \textbf{0.432} & \textbf{0.434} & 0.420 & 0.330 & 0.338 \\ \hline
Intermediate+Target & 0.474 & 0.485 & 0.425 & 0.423 & \textbf{0.422} & \textbf{0.341} & \textbf{0.351} \\ \hline \hline
\end{tabular}}
\label{table:combination}
\end{table*}

\newpage

\subsection{Qualitative Visualization of Performance Improvement.}
\label{sec:quali}

\begin{figure*}[b]
	\centering
	\vspace{-5mm}
	\includegraphics[width=6.8in]{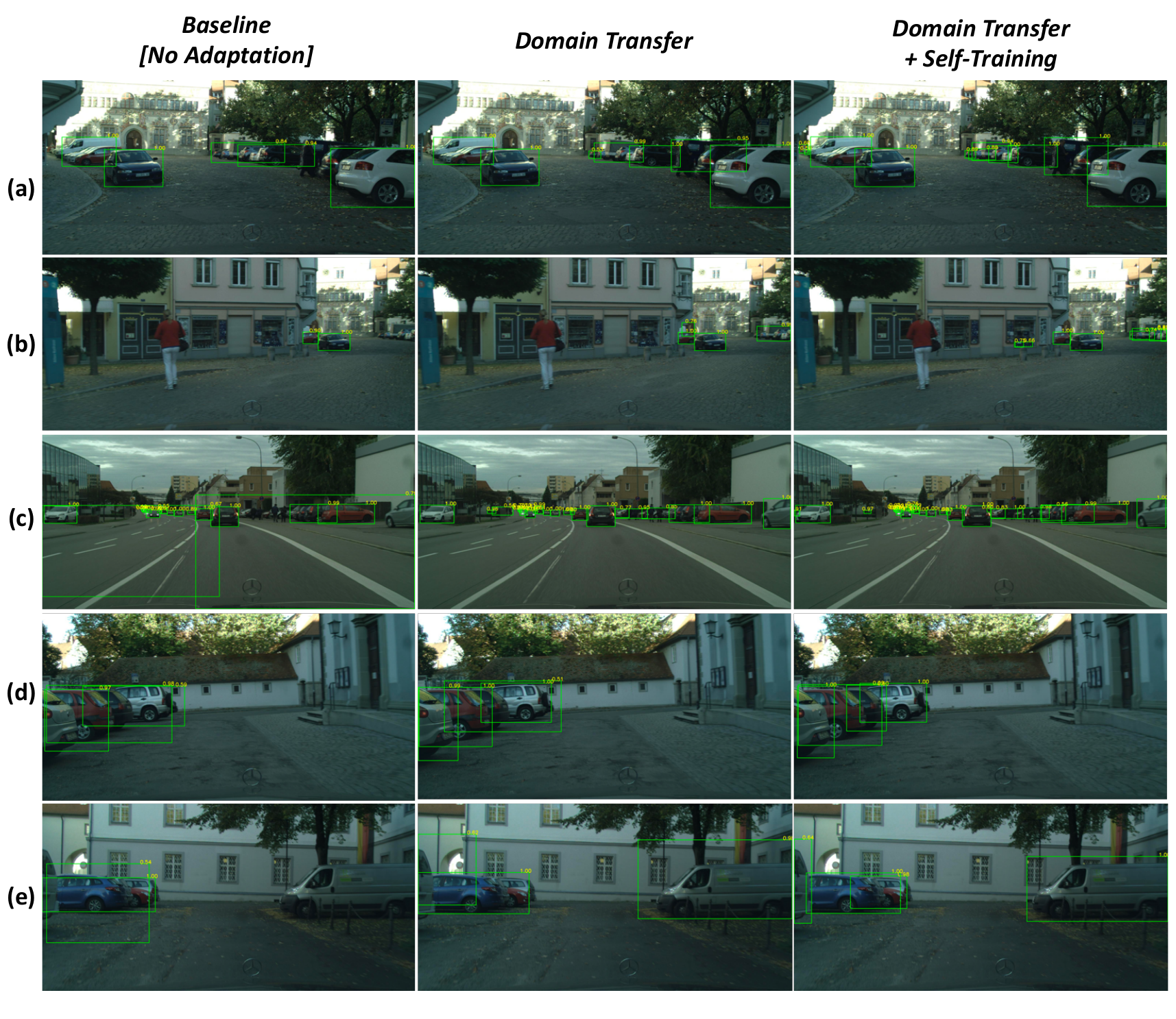}
	\vspace{-3mm}
	\caption{Sim2City Qualitative Visualization. 
	(a)-(c) demonstrates one advantage of our proposed framework is to handle the \textbf{densely-overlapped car scenarios}, which is challenging for non-adapted model. 
	(d)-(e) In addition to that, self-training could also help model to predict \textbf{more accurate box coordinates} (boxes are more tight to the objects themselves.)}
	\label{fig:sim2city_vis}
\end{figure*}

In this section, we qualitatively analyze the domain adaptation performance improvement in this section. 
	Fig.~\ref{fig:sim2city_vis} shows several examples demonstrating the detection performance improvements brought by our fine-grained domain transfer and iterative self-training. 
	The evaluated dataset pair is SIM10K to Cityscapes using MASKRCNN with ResNet50 backbone.
	We can mainly observe two advantages of our adaptation framework in detection performance improvement:
\begin{itemize}
	\item The adapted model is better at recognizing highly-overlapped cars, as shown in Fig.~\ref{fig:sim2city_vis}(a)-(c). 
	This implies that our adapted models learn better \textit{car} feature representation, which is potentially benefited from both domain style transfer and involving target-domain images in self-training.
	
	\item The adapted model can also predict more accurate box coordinates, as shown in Fig.~\ref{fig:sim2city_vis}(d)-(e). 
	As we discussed earlier, this might be because we combine the source-domain GT labels' regression supervision in the self-training process. 
	Even though the pseudo labels may be inaccurate, the model can still receive good regression supervision from source GT labels.
\end{itemize}

\newpage

\subsection{Quantitative Visualization of Performance Improvement.}
\label{sec:quati}

In this section, we quantitatively demonstrate the the pseudo label quality improvement after the first iteration of self-training. 
	To do so, we use the ground-truth (GT) labels and calculate the pseudo labels' IoU with GT labels to show their IoU distribution.
	Note that, this is only for analysis purpose and we did NOT use any GT labels during the self-training process.
	
\begin{figure*}[b]
	\centering
	\vspace{-7mm}
	\includegraphics[width=6.7in]{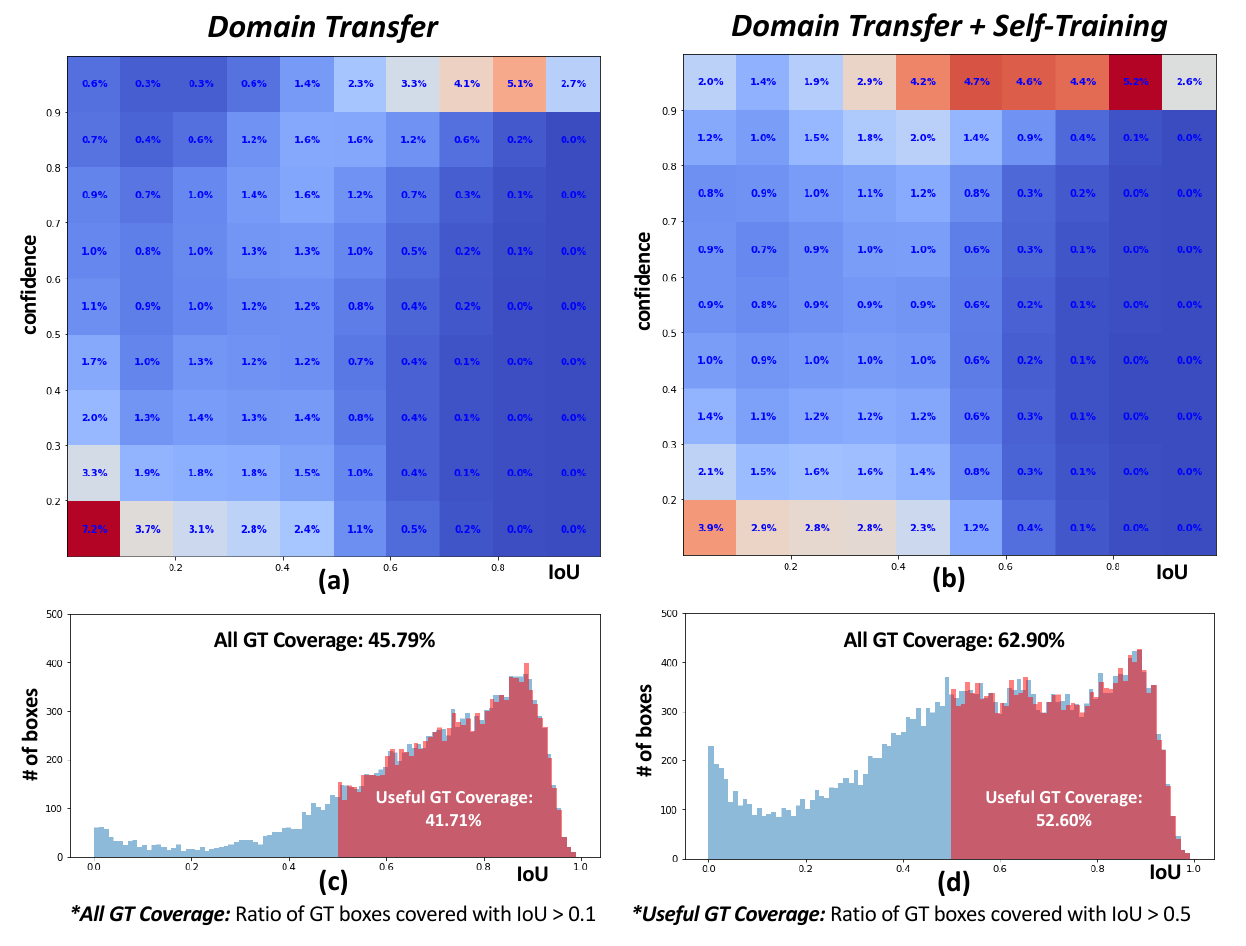}
	\vspace{-3mm}
	\caption{Pseudo Label Quality Improvement Analysis on SIM10K to Cityscapes adaptation. 
	\textbf{Top:} The IoU-Confidence distribution comparison between (a)-(b) demonstrates that the self-training improves the pseudo label quality since more high-IOU boxes are generated (the top right region). 
	\textbf{Bottom:} The comparison of GT box coverage ratios (c)-(d) also shows that after self-training, the model prediction can cover more ground-truth boxes, \textit{e.g.}, from 41.71\% to 52.60\%.}
	\label{fig:quanti_improve}
\end{figure*}

Fig.~\ref{fig:quanti_improve}~(a)-(b) shows the pseudo labels' IoU-Confidence distribution maps on Cityscapes training dataset. 
	The confidence ($y$-axis) is the prediction confidence of each box, and the IoU ($x$-axis) is the highest intersection-over-union of this box with any ground-truth box.
	Comparing the pseudo label distribution between DT model (Fig.~\ref{fig:quanti_improve} (a)) and DT+ST model (Fig.~\ref{fig:quanti_improve} (b))), we can see that \textbf{the number of high-quality boxes significantly increases (the upper right corner)}, demonstrating the performance improvement brought by the self-training.

In the Fig.~\ref{fig:quanti_improve}~(c)-(d), we show the pseudo label's IoU distribution above the confidence threshold (0.9). 
	First, we can see that after self-training, the quantity of high-quality pseudo labels increase a lot (iou$>$0.5).
	We also calculate the \textit{ground-truth (GT) coverage}: the percentage of GT boxes that are covered by the pseudo labels, which can roughly reflect how many objects are detected. 
	Clearly, we can see that self-training improves the \textit{useful GT coverage ratio} (IoU$>0.5$) from 41.7\% to 52.6\% (+10.9\%).
	This means that \textbf{self-training helps the model detect more objects than the previous model}.

Besides, Fig.~\ref{fig:quanti_improve} also quantitatively demonstrates the \textit{error accumulation} phenomenon as we mentioned before.
	Comparing DT and DT+ST model's IoU distributions (a) and (b), we can see that with the increase of high-quality pseudo labels, the low-quality ones (the upper left part) also increase.
	Therefore, we conduct imbalanced sampling and progressively increase the confidence threshold to avoid including too many label errors in the pseudo labels.

\end{document}